\documentclass[conference]{IEEEtran}
\IEEEoverridecommandlockouts                            

\usepackage{cite}
\usepackage{amsmath,amssymb,amsfonts}
\usepackage{graphicx}
\usepackage{textcomp}
\usepackage{xcolor}
\def\BibTeX{{\rm B\kern-.05em{\sc i\kern-.025em b}\kern-.08em
    T\kern-.1667em\lower.7ex\hbox{E}\kern-.125emX}}
\usepackage{amsmath,amsfonts,amssymb}
\usepackage{multirow}
\usepackage{algorithm}
\usepackage{array}
\usepackage{bm}
\usepackage{algpseudocode}
\usepackage{algpseudocodex}

\usepackage{textcomp}
\usepackage{stfloats}
\usepackage{authblk}
\usepackage{url}
\usepackage{subfigure}
\usepackage{verbatim}
\usepackage{graphicx}
\graphicspath{ {./images/} }
\usepackage{cite}
\usepackage{subeqnarray}
\usepackage{cases}
\usepackage{float}
\usepackage{booktabs}
\makeatletter
\let\NAT@parse\undefined
\makeatother
\usepackage{hyperref}
\hypersetup{hypertex=true,
            colorlinks=true,
            urlcolor=black,
            linkcolor=blue,
            anchorcolor=blue,
            citecolor=blue}
\usepackage{cleveref} 

\title{Trajectory Planning and Tracking of Hybrid Flying-Crawling Quadrotors\\
\thanks{$^{1}$The authors are with the Key Laboratory of Smart Manufacturing in Energy Chemical Process Ministry of Education, East China University of Science and Technology, Shanghai, 200237, China.}%
\thanks{$^{2}$The author is with the Research Institute of Intelligent Complex Systems, Fudan University, Shanghai, 200433, China.}%
\thanks{*Corresponding author’s e-mail: yangtang@ecust.edu.cn (Y.Tang)}
}

\author{Dongnan Hu$^{1}$, Ruihao Xia$^{1}$, Xin Jin$^{2}$, and Yang Tang$^{1*}$, \textit{Fellow}, \textit{IEEE}
}

\begin{document}



\maketitle

\begin{abstract}

Hybrid Flying-Crawling Quadrotors (HyFCQs) are transformable robots with the ability of terrestrial and aerial hybrid motion. This article presents a trajectory planning and tracking framework designed for HyFCQs. In this framework, a terrestrial-aerial path-searching method with the crawling limitation of HyFCQs is proposed to guarantee the dynamical feasibility of trajectories. Additionally, a trajectory tracking method is proposed to address the challenges associated with the deformation time required by HyFCQs, which makes tracking hybrid trajectories at the junction between terrestrial and aerial segments difficult. Simulations and real-world experiments in diverse scenarios validate the exceptional performance of the proposed approach.

\end{abstract}

\begin{IEEEkeywords}
trajectory planning and tracking, autonomous navigation
\end{IEEEkeywords}

\section{INTRODUCTION}

In recent years, Unmanned Aerial Vehicles (UAVs) play a significant role in various applications, such as disaster inspection \cite{c_rescue} and transportation \cite{c_delivery} because of their high flexibility and mobility \cite{c_mobility}. Compared with Unmanned Ground Vehicles (UGVs), UAVs lack the abilities, such as high power efficiency and robustness, which limits their applications \cite{c_applications}. To combine UAVs’ high mobility with UGVs’ energy efficiency, several researchers pay attention to terrestrial-aerial robots \cite{c_ter_aer_rob}. 

Hybrid Flying-Crawling Quadrotors (HyFCQs) \cite{c6,c7} represent a category of terrestrial-aerial robots with the ability to transition between different motion modes through morphing mechanisms. This capability enables them to utilize advantages from both terrestrial and aerial mobility. HyFCQs possess characteristics such as compact design, stable propulsion, and adjustable dimensions in crawling mode. These features make them well-suited for navigating through constrained environments, including caves, pipelines, and narrow tunnels, enabling them as ideal robots for autonomous navigation in restricted spaces \cite{c6}.


\begin{figure}[!t]
    \centering
    \setlength{\abovecaptionskip}{0.0cm}
    \includegraphics[width=0.48\textwidth]{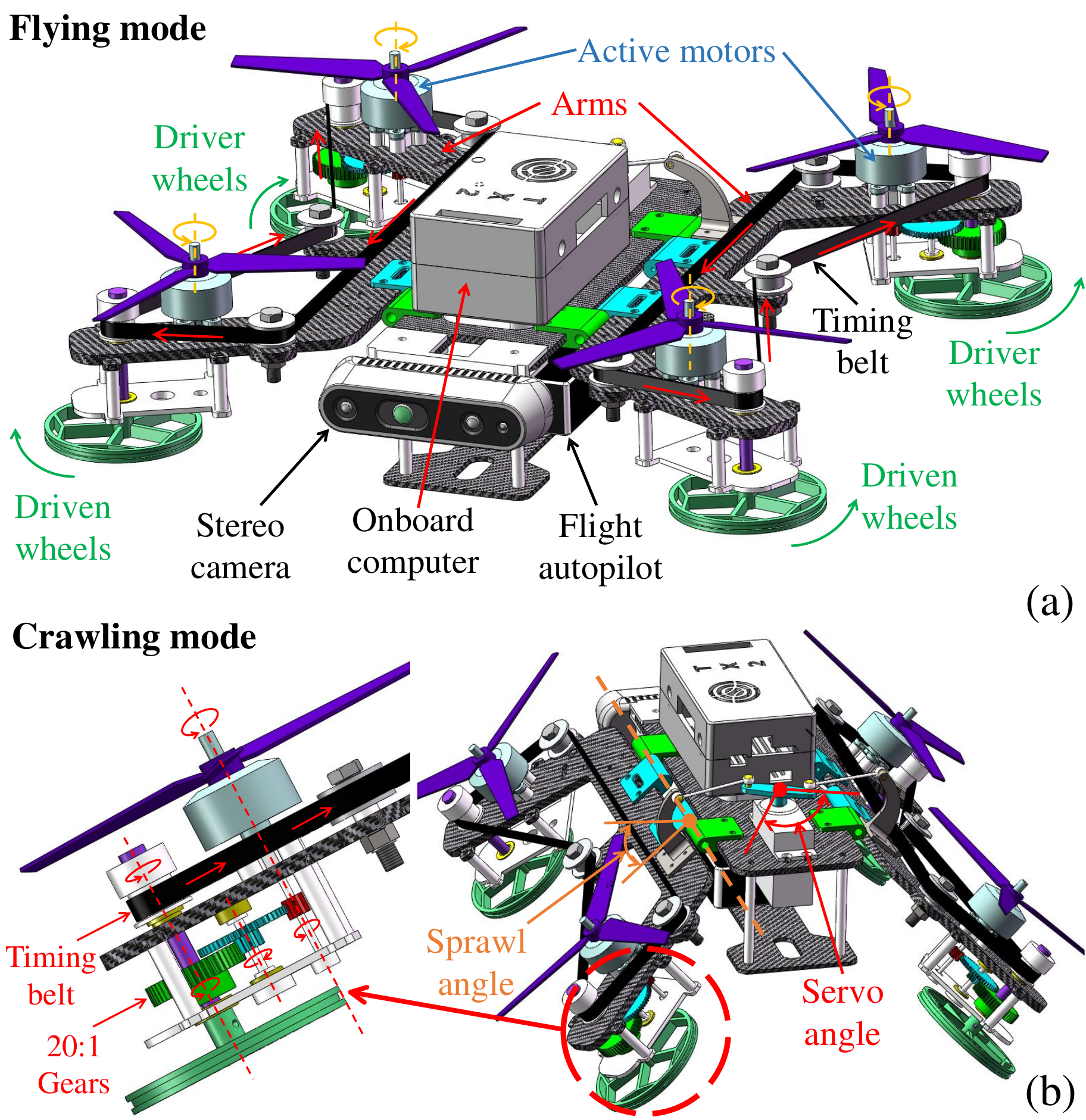}\\
    \caption{Mechanical structure diagram of the HyFCQ. (a) The body core is equipped with sensing, computing, and control units. The arms are equipped with actuation components consisting of reduction gears, timing belts, brushless motors, and wheels. They are linked to the servo mounted in the core. (b) The reduction gears transfer the torsion from the active motors to the rear wheels. The rear wheels convey the torque to the front wheels via timing belts. Symmetrical actuation of the sprawl angles on both sides is achieved through the servo. The experimental video is available at: \href{https://youtu.be/nxFqLxel4c0}{https://youtu.be/nxFqLxel4c0}}
    \label{Fig_solidworks}
    \vspace{-0.5cm}
\end{figure}

Autonomous navigation is widely applied in robots for moving safely in complex environments \cite{c8,c9}. The trajectory planning and tracking modules are crucial for achieving autonomy in HyFCQs. However, existing research mainly focuses on innovations in mechanical structures \cite{c10}, with less attention on the trajectory planning and tracking of HyFCQs. Several challenges of autonomous navigation for such quadrotors need to be addressed, including:
\begin{enumerate}
\item{The HyFCQ relies on a flight autopilot to control the motor speeds for crawling, while the speed of each motor cannot be adjusted independently and freely. This limitation restricts the minimum crawling turning radius of the quadrotor. Therefore, trajectories that involve a movement direction significantly deviating from the current yaw angle of the HyFCQ are infeasible for its crawling motion.}
\item{During trajectory tracking, the system sends the setpoint corresponding to the current timestamp on the trajectory as the desired state to the trajectory tracking module. Therefore, this desired state progressively moves towards the end of the trajectory over time. As the HyFCQs transition between motion modes, structural deformation occurs, necessitating a processing duration. Meanwhile, the setpoint continues advancing towards the trajectory's end according to timestamps. Consequently, by the time the quadrotor completes its deformation and is ready to execute a new motion mode, the setpoint may have moved significantly from the quadrotor's current position. This deviation makes it challenging for the quadrotor to smoothly track the trajectory during transitions between terrestrial and aerial phases.}
\end{enumerate}



In this work, a trajectory planning and tracking framework is designed to address the challenges posed by the structural and motion constraints of HyFCQs, enabling the quadrotors to move through complex environments autonomously. Firstly, the terrestrial-aerial path-searching algorithm with crawling refinement ensures the dynamic feasibility of trajectories. Then we design a terrestrial-aerial trajectory tracking algorithm responsible for terrestrial tracking control, aerial tracking control, and autonomous transition. 

Our contributions can be summarized as follows:
\begin{enumerate}
\item{A terrestrial-aerial trajectory planner is proposed. Compared with existing terrestrial-aerial planner \cite{c11}, it takes into the crawling limitations of HyFCQs in terrestrial path-searching, thus enabling the generation of dynamically feasible hybrid trajectories.}
\item{A terrestrial-aerial trajectory tracking method is proposed. This method achieves autonomous locomotion by avoiding the tracking of terrestrial-aerial junctions and re-planning the trajectory, compensating for the disadvantage of the extended deformation time required by HyFCQs. Additionally, we remap the control inputs originally used for flying motion to suit crawling motion. We then design a ground trajectory tracking controller and integrate it with the existing aerial trajectory tracking controller \cite{c14}, achieving layered terrestrial-aerial trajectory tracking.}
\item{Planner comparisons show the advantage of the proposed algorithm in trajectory feasibility. Extensive simulations and real-world experiments in different restricted scenarios effectively verify the effectiveness of the proposed framework.}
\end{enumerate}

\section{RELATED WORK}

\subsection{Terrestrial-Aerial Trajectory Planning}

There are methods that exist for terrestrial-aerial trajectory planning of passive-wheeled quadrotors \cite{c11,c12}. Fan \emph{et al} \cite{c12} utilized the hybrid A* method, adding energy costs to aerial nodes to prioritize terrestrial paths. Building on this, Zhang \emph{et al}. \cite{c11} applied kinodynamic searching and nonlinear optimization to refine trajectories,  ensuring smooth navigation and avoiding trajectories with excessive curvature changes on the ground. 

However, the above planners are not compatible with HyFCQs: the terrestrial search process does not account for the crawling constraints of HyFCQs, leading to ground trajectories that deviate significantly from the quadrotor's current yaw angle, rendering these trajectories dynamically infeasible. 

\subsection{Terrestrial-Aerial Trajectory Tracking}
Several works proposed trajectory tracking controllers for passive-wheel quadrotors\cite{c10,c11}. Zhang \emph{et al}. \cite{c11} introduced an adaptive thrust control method for terrestrial motion and integrated the nonlinear controller \cite{c14} for aerial trajectory tracking. Building upon the principles of differential flatness, Pan \emph{et al}. \cite{c10} developed a high-speed controller that amalgamates the dynamics of ground support and friction forces.

Unlike passive-wheeled quadrotors that can switch motion modes instantaneously, the transition of motion modes in HyFCQs requires some processing time, making it challenging to track the terrestrial and aerial junction part of the trajectory.


\section{BRIEF OVERVIEW OF THE QUADROTOR}
\label{platform_overview}
The proposed HyFCQ's structure is based on Fcstar \cite{c7}, with its principles illustrated in Fig. \labelcref{Fig_solidworks}. It contains two primary components: the body core and the arms. The body core contains various electronic components such as flight autopilot, onboard computer, battery, stereo camera, and servo. The arms are distributed on both sides and equipped with brushless motors, propellers, reduction gears, timing belt mechanisms, and wheels. They are connected to the servos on the body core through ball joint linkage mechanisms. Therefore, HyFCQs can switch between motion modes by the rotation of the servo motor.

HyFCQ has two modes of motion: flying mode and crawling mode. In flying mode, similar to conventional quadrotors, HyFCQ achieves propulsion through the rotation of the propellers driven by motors. In crawling mode, propulsion is facilitated through the rear active motors. As the motors rotate, they drive the shafts beneath them to rotate and transfer torque to the rear wheels through the gears with a reduction ratio of 20:1. The torque from the rear wheels is transmitted to the front wheels via timing belt mechanisms, thereby achieving four-wheel drive.

\begin{algorithm}
\caption{Terrestrial-Aerial Path Searching}
\label{path-searching}
    \begin{algorithmic}[1]
    \State Initialize();
    \While{$\neg P.\textbf{empty}()$}
        \State $n_c \gets P.\textbf{pop}(), C.\textbf{insert}(n_c)$;
        \If{\textbf{ReachGoal}($n_c$) $\lor$ \textbf{AnalyticExpand}($n_c$)}
            \State \textbf{return} \textbf{RetrievePath}();
        \EndIf
        \If{$n_c.z \geq z_{\textit{threshold}}$}
            \State $primitives \gets \textbf{AerialExpand}(n_c)$;
        \Else
            \State $primitives \gets \textbf{TerrestrialExpand}(n_c)$;
        \EndIf
        \State $nodes \gets \textbf{Prune}(primitives)$;
        \For{$n_i$ in $nodes$}
            \If{$\neg C.\textbf{contain}(n_i) \land \textbf{CheckFeasible}(n_i)$}
                \State $g_{\textit{temp}} \gets n_i.g_{\textit{air}} + n_c.g_c + \textbf{EdgeCost}(n_i)$;
                \If{$\neg P.\textbf{contain}(n_i)$}
                    \State $P.\textbf{add}(n_i)$;
                \ElsIf{$g_{\textit{temp}} \geq n_i.g_c$}
                    \State \textbf{continue};
                \EndIf
                \State $n_i.\textit{parent} \gets n_c, n_i.g_c \gets g_{\textit{temp}}$;
                \State $n_i.f_c \gets n_i.g_c + \textbf{Heuristic}(n_i)$;
            \EndIf
        \EndFor
    \EndWhile
    \end{algorithmic}
\end{algorithm}

\section{TERRESTRIAL-AERIAL TRAJECTORY PLANNING}

The proposed trajectory planner is based on the Fast-Planner framework \cite{c8} which encompasses hybrid A* based path-searching for the front end and trajectory optimization for the back end. To enhance the dynamic viability of the ground portion of the hybrid trajectory, the path-searching module incorporates yaw constraints from the quadrotor during the expansion of terrestrial primitives.

The pseudo-code of proposed path-searching is illustrated in Alg. \ref{path-searching}, where \(P\) and \(C\) denote the open and closed set respectively. The variable \(g_c\) refers to the actual cost of the path from the start position to the current node, and \(g_{air}\) represents the aerial cost, set to a positive constant for nodes above a certain altitude and zero below, directing the planner towards ground paths. Each node \(n\) records attributes such as a primitive, the ending voxel of the primitive, the center position of the ending voxel, \(g_c\), \(g_{air}\) and \(f_c\). \textbf{AnalyticExpand}() computes an optimal path from the current node to the goal, ending the search in advance if the path is collision-free and dynamically feasible. Primitives are expanded iteratively and pruned (\textbf{Prune}()), retaining the least cost \(f_c\). \textbf{CheckFeasible}() evaluates the safety and feasibility of remaining primitives, retaining the node with the lowest \(f_c\). This process continues until any primitive reaches the goal or the \textbf{AnalyticExpand}() succeeds. Details of \textbf{AnalyticExpand}(), \textbf{CheckFeasible}(), \textbf{EdgeCost}() and \textbf{Heuristic}() are in \cite{c8}, while this section elaborates on \textbf{TerrestrialExpand}() and \textbf{AerialExpand}(), which correspond to the generation of terrestrial and aerial primitives, respectively.

After path-searching, the entire path needs further optimization to generate a smooth and feasible trajectory, which benefits the quadrotors' tracking. The optimization method is illustrated in \cite{c11}.

\begin{figure}[t]
    \centering
    \setlength{\abovecaptionskip}{0.0cm}
    \includegraphics[width=0.48\textwidth]{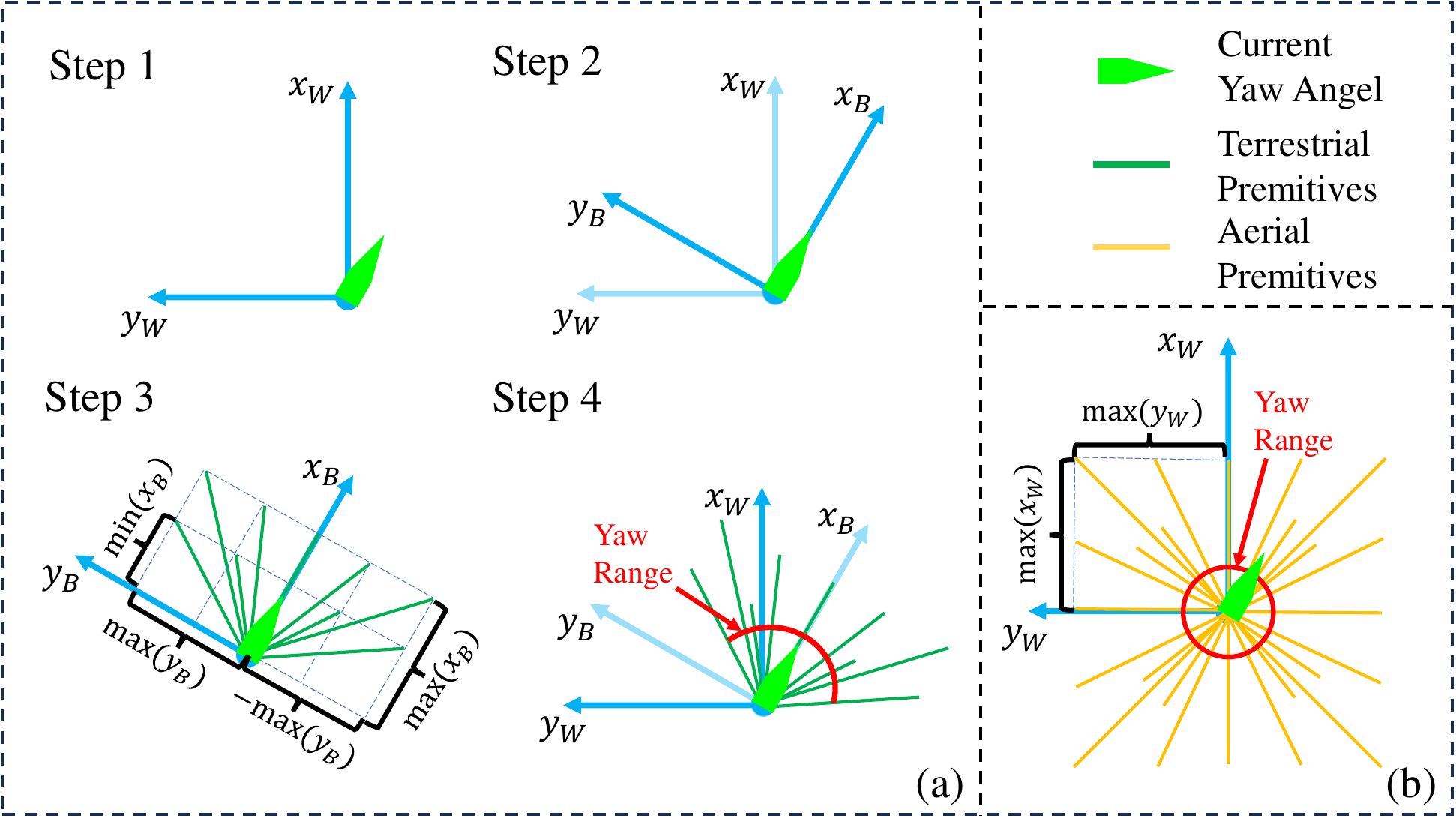}\\
    \caption{Top-down view of the primitives expansion. (a) shows the process of terrestrial primitives expansion. By increasing the minimum value of \(x_B\), the variation range of the yaw angle is restricted, thereby reducing the fluctuation of the yaw angle. (b) shows the process of aerial primitives expansion. Since the flying motion is unaffected by nonholonomic constraints, the primitives can expand in all directions.}
    \label{Fig_primitives_expansion}
    \vspace{-0.5cm}
\end{figure} 

\subsection{Terrestrial Primitives Generation}





We first discuss \textbf{TerrestrialExpand}(), the terrestrial primitives generation function. The visualization of this process is shown in Fig. \ref{Fig_primitives_expansion}(a). During crawling locomotion, due to the nonholonomic constraint and the limitations of the turning radius, it is necessary to maintain a relatively small disparity between the yaw angle of the trajectory and the present yaw angle of the HyFCQ. Therefore, during the generation of trajectory primitives, we apply constraints to the trajectory with respect to the current yaw angle, which is denoted as \(\psi _{t_0}^{W}\). The state space model for terrestrial search can be defined as follows:
\begin{subequations}\label{eq16}
    \begin{align}
    \mathbf{\dot{x}}_{t}^{W} &= \mathbf{R}_{d}^{-1}\mathbf{AR}_d\mathbf{x}_{t}^{W} + \mathbf{R}_{d}^{-1}\mathbf{Bu}_{t}^{B}, \\
    \mathbf{R}_d &= \left[ \begin{matrix}
    \mathbf{R}& \bm{0}\\
    \bm{0}& \mathbf{R}\\
    \end{matrix} \right] ,\mathbf{A} = \left[ \begin{matrix}
    \bm{0}& \mathbf{I}_3\\
    \bm{0}& \bm{0}\\
    \end{matrix} \right] ,\mathbf{B} = \left[ \begin{array}{c}
    \bm{0}\\
    \mathbf{I}_3\\
    \end{array} \right],  \label{eq16B}\\
    \mathbf{R} &= \left[ \begin{matrix}
    \cos \left( \psi _{t_0}^{W} \right)& \sin \left( \psi _{t_0}^{W} \right)& 0\\
    -\sin \left( \psi _{t_0}^{W} \right)& \cos \left( \psi _{t_0}^{W} \right)& 0\\
    0& 0& 1\\
    \end{matrix} \right],  \label{eq16C}\\
    \psi _{t}^{W} &= \tan ^{-1}\frac{\left( \mathbf{p}_{t}^{W} \right) _y-\left( \mathbf{p}_{t_0}^{W} \right) _y}{\left( \mathbf{p}_{t}^{W} \right) _x-\left( \mathbf{p}_{t_0}^{W} \right) _x}, \label{eq16D}
    \end{align}
\end{subequations}
where \(\mathbf{x}_{t}^{W}=\left[ \left( \mathbf{p}_{t}^{W} \right) ^{\text{T}},\left( \mathbf{\dot{p}}_{t}^{W} \right) ^{\text{T}} \right] ^{\text{T}}\in \mathbb{R}^6\), denoted as the state of the system in world coordinate. \(\mathbf{p}_{t}^{W}=\left[ \left( \mathbf{p}_{t}^{W} \right) _x,\left( \mathbf{p}_{t}^{W} \right) _y,\left( \mathbf{p}_{t}^{W} \right) _z \right] ^{\text{T}}\), denoted as the position at time t. \(\left( \mathbf{p}_{t}^{W} \right) _x\), \(\left( \mathbf{p}_{t}^{W} \right) _y\) and \(\left( \mathbf{p}_{t}^{W} \right) _z\) represent the components of \(\mathbf{p}_{t}^{W}\) along x-y-z axis. \(\mathbf{R}\) is the rotation matrix based on the initial state's yaw \(\psi _{t_0}^{W}\) which converts state variables from the world coordinate to the body coordinate. \(\mathbf{u}_{t}^{B}=\left[ \left( \mathbf{u}_{t}^{B} \right) _x,\left( \mathbf{u}_{t}^{B} \right) _y,\left( \mathbf{u}_{t}^{B} \right) _z \right] ^{\text{T}}\) is control input in the body coordinate, representing acceleration. Then, the state transition result in the body coordinate is calculated based on the input \(\mathbf{u}_{t}^{B}\). Finally, the result is transformed into the world coordinate by premultiplying \(\mathbf{R}_{d}^{-1}\). The solution for the state equation can be expressed as:
\begin{equation}
\mathbf{x}_{t}^{W}=e^{\mathbf{R}_{d}^{-1}\mathbf{AR}_dt}\mathbf{x}_{t_0}^{W}+\int_0^t{e^{\mathbf{R}_{d}^{-1}\mathbf{AR}_d\left( t-\tau \right)}\mathbf{R}_{d}^{-1}\mathbf{Bu}_{\tau}^{B}d\tau}.
\end{equation}
Each dimension of the acceleration input is discretized by resolution \(r\), while \(u_{\max}\) is determined by user-defined settings:
\begin{subequations}\label{eq16}
    \begin{align}
    \left( \mathbf{u}_{t}^{B} \right) _x\in \left\{ \left( \frac{\alpha}{r} \right) u_{\max},\left( \frac{\alpha +1}{r} \right) u_{\max},...,u_{\max} \right\} ,  \label{}\\
    \left( \mathbf{u}_{t}^{B} \right) _{y,z}\in \left\{ -u_{\max},-\frac{r-1}{r}u_{\max},...,\frac{r-1}{r}u_{\max},...,u_{\max} \right\} . \label{}
    \end{align}
\end{subequations}
Where \(\alpha\) is an integer variable ranging from 0 to \(r\) and is used to limit the minimum \(\left( \mathbf{u}_{t}^{B} \right) _x\). Increasing the lower limit of \(\left( \mathbf{u}_{t}^{B} \right) _x\) can increase the minimum value of \(x_B\), as illustrated in Fig. \ref{Fig_primitives_expansion}(a), thereby limiting the yaw angle range. Consequently, when \(\alpha\) is increased, the maximum yaw deviation decreases. However, an excessively large value for \(\alpha\) may result in path-searching failures. The appropriate value of \(\alpha\) should be determined through experimental analysis. Then the calculated yaw angle \(\psi _{t}^{W}\) from this step is utilized as the initial yaw angle \(\psi _{t_0}^{W}\) for the next step in primitive generation.

\subsection{Aerial Primitives Generation}

If the height of the current node \(n_c.z\) exceeds or meets the threshold, indicating that the node is airborne, the primitives' expansion method is aerial primitives generation \textbf{AerialExpand}(), as shown in Fig. \ref{Fig_primitives_expansion}(b). In this scenario, the quadrotor is free from non-holonomic constraints. Consequently, the state space model for aerial search can be defined as follows:
\begin{subequations}\label{eq17}
    \begin{align}
    \mathbf{\dot{x}}_{t}^{W}=\mathbf{Ax}_{t}^{W}+\mathbf{Bu}_{t}^{W},
    \label{eq17B}\\
    \mathbf{A}=\left[ \begin{matrix}
	\bm{0}&		\mathbf{I}_3\\
	\bm{0}&		\bm{0}\\
    \end{matrix} \right] ,\mathbf{B}=\left[ \begin{array}{c}
	\bm{0}\\
	\mathbf{I}_3\\
    \end{array} \right] ,
    \label{eq17D}
    \end{align}
\end{subequations}

The complete solution for the state equation is expressed as:

\vspace{-0.1cm}
\begin{equation}
\mathbf{x}_{t}^{W}=e^{\mathbf{A}t}\mathbf{x}_{t_0}^{W}+\int_0^t{e^{\mathbf{A}\left( t-\tau \right)}\mathbf{Bu}_{\tau}^{W}d\tau}.
\end{equation}
\vspace{-0.15cm}

{
\setlength{\parindent}{0cm}
where \(\mathbf{u}_{t}^{W}\) denoted as the acceleration input in the world coordinate. Each dimension \(\left[ -u_{\max},u_{\max} \right] \) is discretized as \(\left\{ -u_{\max},-\frac{r-1}{r}u_{\max},...,\frac{r-1}{r}u_{\max},...,u_{\max} \right\} \), where \(r\) represents resolution. 
}


\begin{figure}[ht]
    \centering
    \setlength{\abovecaptionskip}{0.0cm}
    \includegraphics[width=0.47\textwidth]{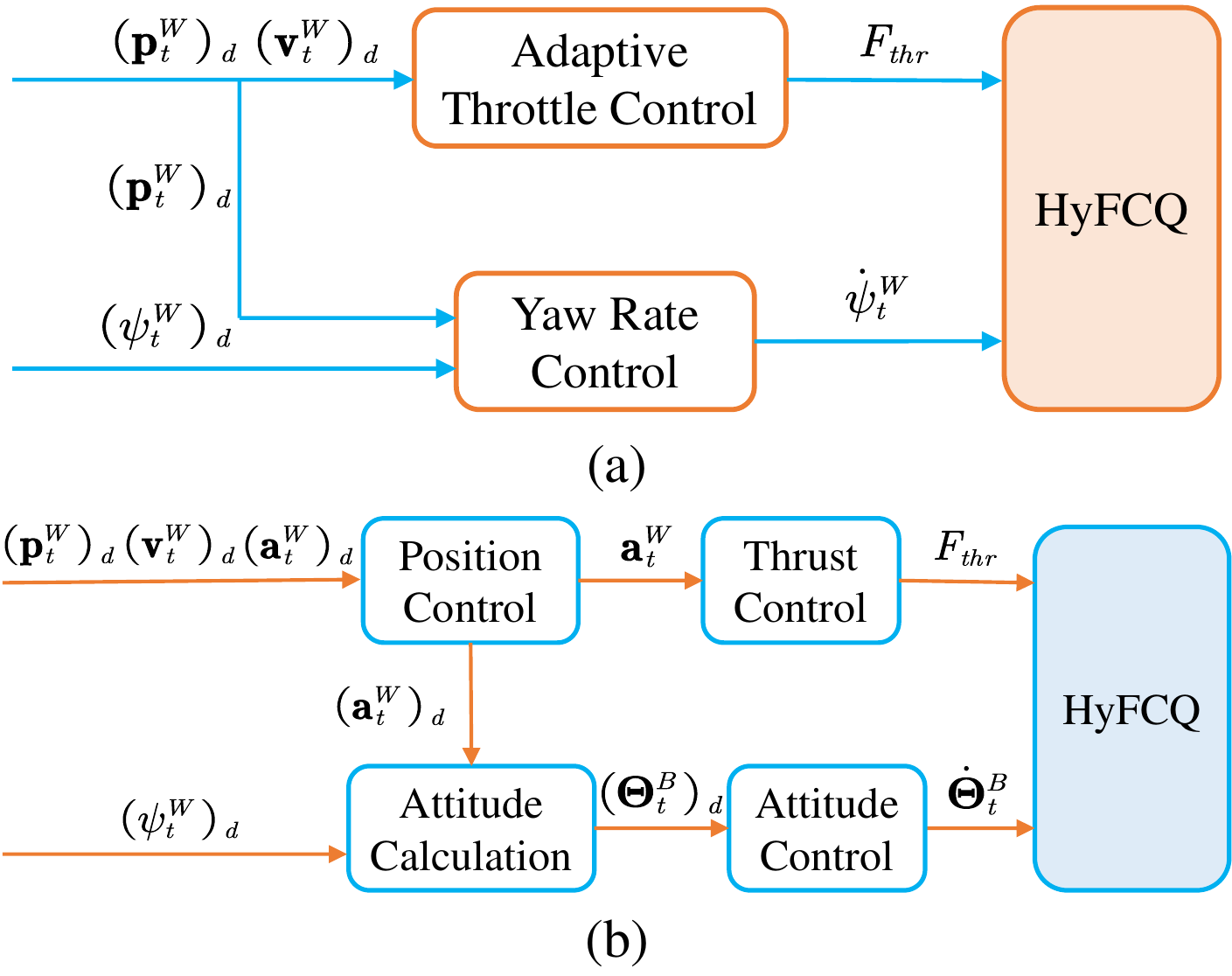}\\
    \caption{(a) The trajectory tracking controller for crawling mode. (b) The trajectory tracking controller for flying mode.}
    \label{Fig_controller}
    \vspace{-0.2cm}
\end{figure} 

\begin{algorithm}
\caption{Terrestrial-Aerial Trajectory Tracking}
\label{trajectory-tracking}
\begin{algorithmic}[1]
\While{$\neg \textbf{ReachGoal}()$}
    \If{$\left( z_{setpoint} \right) _{i} < z_{threshold} \And \left( z_{setpoint} \right) _{i+1} \geq z_{threshold}$}
        \State \textbf{CrawlToFly}()
        \State \textbf{TakeOff}()
        \State \textbf{TrajectoryReplan}()
    \ElsIf{$ \left( z_{setpoint} \right) _{i} \geq z_{threshold} \And \left( z_{setpoint} \right) _{i+1} < z_{threshold}$}
        \State \textbf{Land}()
        \State \textbf{FlyToCrawl}()
        \State \textbf{TrajectoryReplan}()
    \ElsIf{$\left( z_{setpoint} \right) _{i} \geq z_{threshold}$}
        \State \textbf{AerialTrajectoryTrack}()
    \Else
        \State \textbf{TerrestrialTrajectoryTrack}()
    \EndIf
\EndWhile
\end{algorithmic}
\end{algorithm}

\section{TERRESTRIAL-AERIAL TRAJECTORY TRACKING}
\label{section_tracking}
After trajectory generation, a setpoint along the trajectory is chosen based on the current timestamp and transmitted as the desired state to the trajectory tracking module. A terrestrial setpoint includes the yaw angle and a 3D position, velocity (\(\left[ \left( \mathbf{p}_{t}^{W} \right) _d,\ \left( \mathbf{v}_{t}^{W} \right) _d,\ \left( \psi _{t}^{W} \right) _d \right] ^{\text{T}}\)) while an aerial setpoint includes the yaw angle and a 3D position, velocity, acceleration (\(\left[ \left( \mathbf{p}_{t}^{W} \right) _d,\ \left( \mathbf{v}_{t}^{W} \right) _d,\ \left( \mathbf{a}_{t}^{W} \right) _d,\ \left( \psi _{t}^{W} \right) _d \right] ^{\text{T}}\)). The current state of HyFCQ is denoted as \(\mathbf{\hat{X}}_{t}^{W}\). \(\mathbf{\hat{X}}_{t}^{W}=\left[ \mathbf{\hat{p}}_{t}^{W},\mathbf{\hat{v}}_{t}^{W},\mathbf{\hat{\Theta}}_{t}^{W},\mathbf{\hat{\dot{\Theta}}}_{t}^{W} \right] ^{\text{T}}\), respectively correspond to position, velocity, orientation, and angular rates. The orientation \(\mathbf{\hat{\Theta}}_{t}^{W}\) and angular rates \(\mathbf{\hat{\dot{\Theta}}}_{t}^{W}\) are parameterized in the \(X-Y-Z\) Euler angle sequence. \(\mathbf{\hat{\Theta}}_{t}^{W}=\left[ \mathbf{\hat{\phi}}_{t}^{W},\mathbf{\hat{\theta}}_{t}^{W},\hat{\psi}_{t}^{W} \right] ^{\text{T}}\), \(\mathbf{\hat{\dot{\Theta}}}_{t}^{W}=\left[ \mathbf{\hat{\dot{\phi}}}_{t}^{W},\mathbf{\hat{\dot{\theta}}}_{t}^{W},\hat{\dot{\psi}}_{t}^{W} \right] ^{\text{T}}\). The trajectory tracking pseudo-code is shown in Alg. \ref{trajectory-tracking}.

This section uses the PX4 open-source firmware\cite{c13} as the autopilot firmware. The PX4 firmware calculates motor rotational speed based on inputs such as throttle thrust \(F_{thr}\) (a normalized value ranging from 0 to 1) and body angular rates \(\mathbf{\dot{\Theta}}_{t}^{B}\). In flying mode, increasing \(F_{thr}\) elevates the rotational speed of all motors, enhancing propulsive force and causing ascent. Additionally, increasing the body angular velocity \(\mathbf{\dot{\Theta}}_{t}^{B}\), such as the yaw rate \(\dot{\psi}_{t}^{W}\), augments the rotational speed of the quadrotor along the z-axis. In crawling mode, increasing \(F_{thr}\) leads to an increase in forward speed, while elevating the yaw rate \(\dot{\psi}_{t}^{W}\) increases the differential speed between active motors, reducing the turning radius. By mapping the results of the same input to different motion modes, the autopilot originally designed for flight can be adapted for crawling motion.




\subsection{Terrestrial Trajectory Tracking}

The terrestrial trajectory tracking controller \textbf{TerrestrialTrajectoryTrack}() is illustrated in Fig. \ref{Fig_controller}(a). In crawling mode, the pivotal challenge is how to control the yaw rate \(\dot{\psi}_{t}^{W}\) and throttle thrust \(F_{thr}\) to reduce the position error along x-axis and y-axis.

\subsubsection{Yaw Rate Control} Assuming a level ground, it follows that the z-axis in both the world coordinate system and the body coordinate system runs in the same direction. In this context, the yaw rate \(\dot{\psi}_{t}^{W}\) in the world coordinate system and \(\dot{\psi}_{t}^{B}\) in the body coordinate system are identical. Considering yaw angle and position errors, the calculation of \(\dot{\psi}_{t}^{W}\) is:
\begin{equation}
\dot{\psi}_{t}^{W}=K_y\left[ \lambda _d\left( \psi _{t}^{W} \right) _{d}^{e}+\lambda _p\left( \psi _{t}^{W} \right) _{p}^{e} \right], 
\end{equation}
where \(K_y\) represents the proportional gain.  \(\left( \psi _{t}^{W} \right) _{d}^{e}\) is the deviation between the desired yaw angle \(\left( \psi _{t}^{W} \right) _d\) and the current yaw angle \(\hat{\psi}_{t}^{W}\): 
\begin{equation}
\left( \psi _{t}^{W} \right) _{d}^{e}=\left( \psi _{t}^{W} \right) _d-\hat{\psi}_{t}^{W},
\end{equation}
\(\left( \psi _{t}^{W} \right) _{p}^{e}\) represents the yaw angle correction needed to eliminate the position error \(\left( \mathbf{p}_{t}^{W} \right) ^e\):
\begin{subequations}\label{eq18}
    \begin{align}
    \left( \psi _{t}^{W} \right) _{p}^{e}=tan^{-1}\frac{\left( \mathbf{p}_{t}^{W} \right) _{y}^{e}}{\left( \mathbf{p}_{t}^{W} \right) _{x}^{e}}-\hat{\psi}_{t}^{W},
    \label{eq17B}\\
    \left( \mathbf{p}_{t}^{W} \right) ^e=\left( \mathbf{p}_{t}^{W} \right) _d-\mathbf{\hat{p}}_{t}^{W},
    \label{eq17D}
    \end{align}
\end{subequations}
where \(\left( \mathbf{p}_{t}^{W} \right) _{x}^{e}\) and \(\left( \mathbf{p}_{t}^{W} \right) _{y}^{e}\) respectively represent the value of \(\left( \mathbf{p}_{t}^{W} \right) ^e\) along x-axis and y-axis. Variables \(\lambda _d\) and \(\lambda _p\) serve as weight parameters, and their values are adjusted based on specific circumstances, with the constraint that \(\lambda _d+\lambda _p=1\). When the norm of \(\left( \mathbf{p}_{t}^{W} \right) ^e\) exceeds a specific threshold, indicating a large positional error, more weight is given to correcting the yaw angle error during the calculation of \(\dot{\psi}_{t}^{W}\), resulting in increased \(\lambda _p\) and reduced \(\lambda _d\). Conversely, when the norm is below this threshold, indicating a minor positional error, the focus shifts towards maintaining the desired yaw, increasing \(\lambda _d\) and decreasing \(\lambda _p\).

\subsubsection{Adaptive Throttle Control} In crawling mode, the forward velocity \(V\) \(\left( m/s \right) \) is determined by the throttle thrust \(F_{thr}\) . With the HyFCQ platform in Sect \ref{results}(A), multiple sets of throttle levels are established and the corresponding forward velocities are recorded. Through a linear regression approach, the relationship between \(V\) and \(F_{thr}\) is:
\begin{equation}
V=6.838F_{thr}+0.0016,  \label{v_f_thr}
\end{equation}
\(V_t\), the forward velocity at time t can be calculated as:

\begin{equation}
V_t=\lVert \left( \mathbf{v}_{t}^{W} \right) _d+K_p\left( \mathbf{p}_{t}^{W} \right) ^e \rVert , \label{forward_velocity}
\end{equation}
where \(\left( v_{t}^{W} \right) _d\) is the desired velocity in the trajectory. \(K_p\) is the proportional gain for the position error. When the position error is substantial, this gain facilitates the rapid convergence of \(V_t\) to the desired value. By combining (\ref{v_f_thr}) with (\ref{forward_velocity}), \(F_{thr}\) can be calculated:
\begin{equation}
F_{thr}=K_{thr}\frac{\left( V_t-0.0016 \right)}{6.838}.
\end{equation}
\(K_{thr}\) is the proportional gain, and increasing its value appropriately can accelerate the response of motion.

In crawling mode, the propellers generate thrust in both the y-axis and z-axis of the quadrotor's body. Excessive thrust can lead to significant lateral deviations and overly rapid crawling speeds. Additionally, when \(F_{thr}\) exceeds 0.28, the linearity between it and \(V\) decreases. In order to avoid the issues mentioned above, we set the maximum value of \(F_{thr}\) in crawling motion as 0.2.


\subsection{Aerial Trajectory Tracking}

In this section, we adopt the existing method in \textbf{AerialTrajectoryTrack}(). The aerial trajectory tracking controller, as shown in Fig. \ref{Fig_controller}(b), includes blocks of position control, thrust control, attitude calculation, and attitude control. The exhaustive equations and explanation of these blocks are in \cite{c14}. The position control block calculates the required acceleration \(\mathbf{a}_{t}^{W}\) based on position and velocity errors with proportional controller, as well as the desired acceleration \(\left( \mathbf{a}_{t}^{W} \right) _d\). Then \(\mathbf{a}_{t}^{W}\) is provided to the thrust control block, which computes the thrust \(F_{thr}\) for the quadrotor. Simultaneously, \(\left( \mathbf{a}_{t}^{W} \right) _d\) and the desired yaw angle \(\left( \psi _{t}^{W} \right) _d\) serve as inputs for the attitude computation module, which calculates the desired angular orientation \(\left( \mathbf{\Theta }_{t}^{B} \right) _d\) in the body frame. Then \(\left( \mathbf{\Theta }_{t}^{B} \right) _d\) is subsequently used as input for the attitude control module. The attitude control module then computes the angular velocities \(\mathbf{\dot{\Theta}}_{t}^{B}\) in the body frame, which are sent to the quadrotor to facilitate the required maneuvers. 


\subsection{Transition between Crawling and Flying}
Due to the deformation required by HyFCQs needing processing time, the quadrotor should avoid tracking the trajectory at the transition between land and air. If the current setpoint altitude \(\left( z_{traj\_setpoint} \right) _{i}\) is below a specific threshold \(z_{threshold}\), and the subsequent timestamp's setpoint altitude \(\left( z_{traj\_setpoint} \right) _{i+1}\) meets or exceeds \(z_{threshold}\), the quadrotor should cease tracking the following trajectory and switch to flying mode (\textbf{CrawlToFly}()). Then HyFCQ will take off to a specific altitude, and hover (\textbf{TakeOff}()), followed by trajectory replanning from its current position to the destination (\textbf{TrajectoryReplan}()). Conversely, if the current setpoint altitude is initially above the threshold and the subsequent setpoint is below it, this signals a switch from air to ground. In this case, the quadrotor should halt tracking and begin landing procedures (\textbf{Land}()), switch to crawling mode, and replan the trajectory to the final goal upon landing. 

Additionally, if the current setpoint altitude \(\left( z_{traj\_setpoint} \right) _{i}\) exceeds or equals the threshold \(z_{threshold}\), it indicates that the quadrotor is flying, and the aerial trajectory tracking controller \textbf{AerialTrajectoryTrack}() is active. Conversely, if the current setpoint altitude is below the threshold, it signifies that the quadrotor is crawling, triggering the terrestrial trajectory tracking controller  \textbf{TerrestrialTrajectoryTrack}(). The process will continue until the quadrotor reaches the goal.

\begin{figure}[t]
    \centering
    \setlength{\abovecaptionskip}{0.0cm}
    \includegraphics[width=0.46\textwidth]{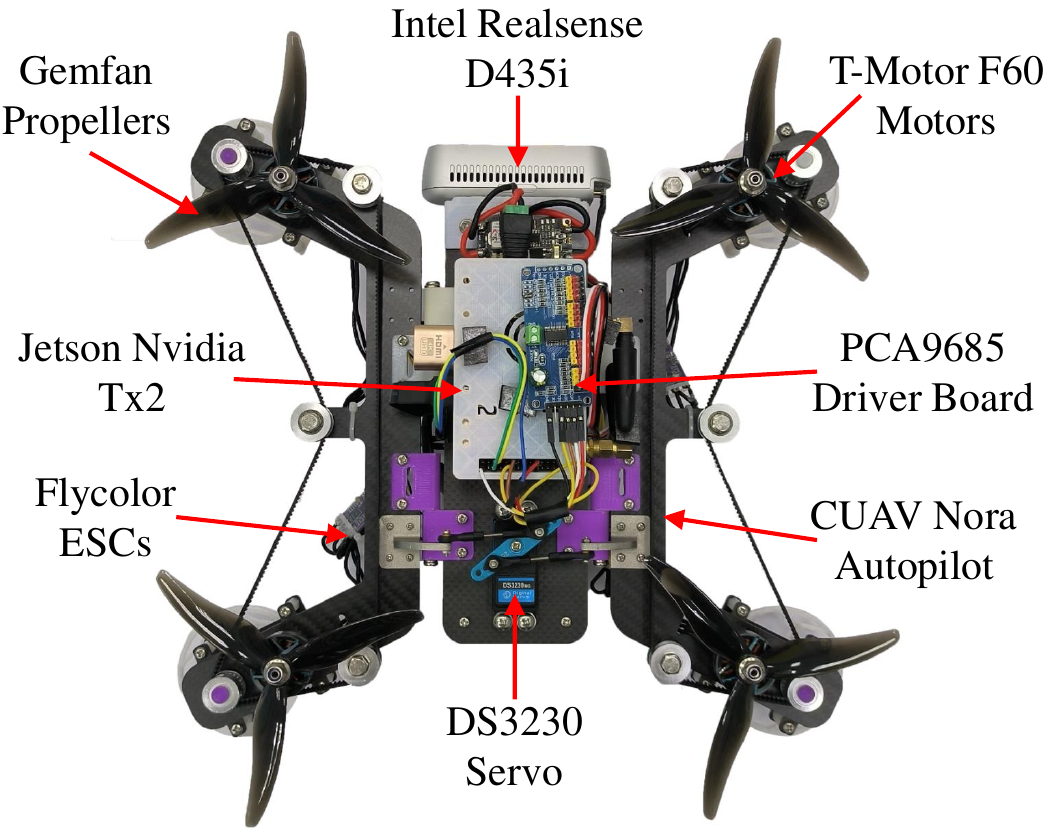}\\
    \caption{The detailed components of the HyFCQ. In flying mode, the size of the quadrotor is \(34\times 32\times 11\) cm, while in crawling mode, the size is \(34\times 26\times 11\) cm. The total weight is \(1.6\) kg.}
    \label{Fig_real_machine}
    \vspace{-0.2cm}
\end{figure} 

\begin{table}[t]
    \renewcommand{\arraystretch}{1.3}
    \caption{Description of The Main Components}
    \label{components}
    \centering
    \begin{tabular}{|c|c|l|}
        \hline
        Components & Part numbers & Key features \\
        \hline
        Onboard computer & Jetson Nvidia TX2 & GPU: 256 CUDA cores \\
        \hline
        Flight autopilot & CUAV Nora & CPU: STM32H743 \\
        \hline
        Stereo camera & Intel Realsense D435i & Range: 0.3-3 \(m\) \\
        \hline
        Motor & T-Motor F60 & 2550KV \\
        \hline
        ESC &  Flycolor 50A & MCU: STM32G071 \\
        \hline
        Propeller & Gemfan 51477 & Diameter: 5 inch \\
        \hline
        Servo & DS3230 & Torque: 320 \(N\cdot cm\) \\
        \hline
        Driver Board & PCA9685 & Resolution: 12-bit \\
        \hline
        Battery & GS33004S30 & Weight: 235 g \\
        \hline
    \end{tabular}
    \vspace{-0.1cm}
\end{table}

\begin{table}[t]
    \renewcommand{\arraystretch}{1.3}
    \caption{The Results Of Terrestrial Trajectory Tracking}
    \label{table_example}
    \centering
    \begin{tabular}{|c|c|c|c|c|}
        \hline
        Trajectory & \(E_{ap} (m)\) & \(E_{mp} (m)\) & \(E_{ay}(rad)\) & \(E_{my}(rad)\) \\
        \hline
        Circular & 0.046 & 0.047 & 0.049 & 0.239 \\
        \hline
        Lemniscate & 0.048 & 0.064 & 0.123 & 0.550 \\
        \hline
    \end{tabular}
    \vspace{-0.2cm}
\end{table}
\begin{figure}[t]
    \vspace{-0.3cm}
    \centering
    \setlength{\abovecaptionskip}{0.0cm}
    \includegraphics[width=0.46\textwidth]{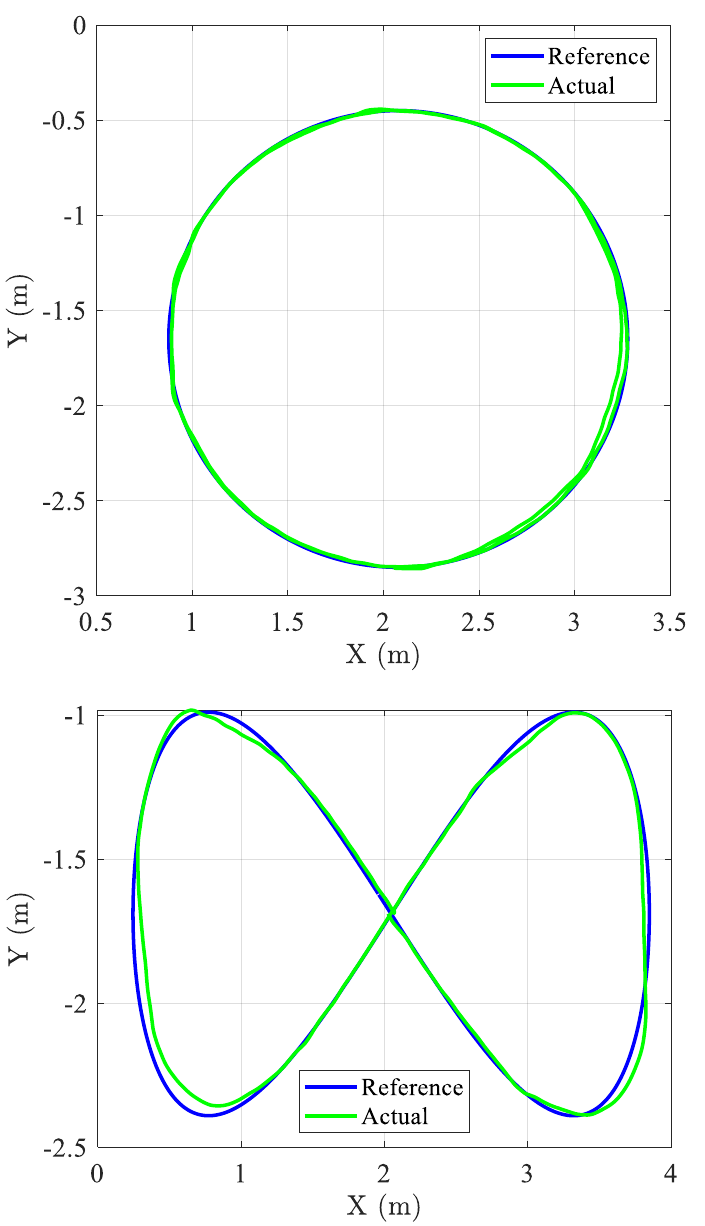}\\
    \caption{Top) The circular trajectory tracking. Botton) The lemniscate trajectory tracking.}
    \label{Fig_8}
    \vspace{-0.5cm}
\end{figure} 

\section{Results}
\label{results}
\subsection{HyFCQ Platform}
\label{HyFCQ_platform}
Based on the mechanical structure of Sect \ref{platform_overview}, the HyFCQ incorporates perception, computation, and actuation modules, as shown in Fig. \ref{Fig_real_machine}. The perception module includes a stereo camera for acquiring depth images. The computation module includes an onboard computer for processing tasks such as localization, mapping, trajectory planning, and tracking. The actuation module consists of an autopilot, electronic speed controllers (ESCs), motors, servo, and servo driver board. Detailed descriptions of the components are provided in Table \ref{components}.

\subsection{Terrestrial Trajectory Tracking}
To validate the effectiveness of the terrestrial trajectory tracking controller, we conduct tests using circular and lemniscate trajectories. The evaluation criteria for trajectory tracking performance include the average position error \(E_{ap} \), the maximum position error \(E_{mp}\), the average yaw angle error \(E_{ay}\) and the maximum yaw angle error \(E_{my}\). The circular has a radius of 1.2 meters, while the lemniscate has a longitudinal length of 3.6 meters and a lateral width of 1.4 meters. The maximum velocity limit for the circular trajectory is 0.8 \(m/s\), and for the lemniscate trajectory, it is 1.0 \(m/s\).

We conduct 10 sets of experiments and performed statistical analysis on the data. The tracking errors and plots are shown in \textbf{Table} \ref{table_example} and Fig. \ref{Fig_8} respectively. It can be seen that the proposed controller achieves stable tracking of the terrestrial trajectories effectively. 


\begin{figure}[t]
    \centering
    \setlength{\abovecaptionskip}{0.0cm}
    \includegraphics[width=0.48\textwidth]{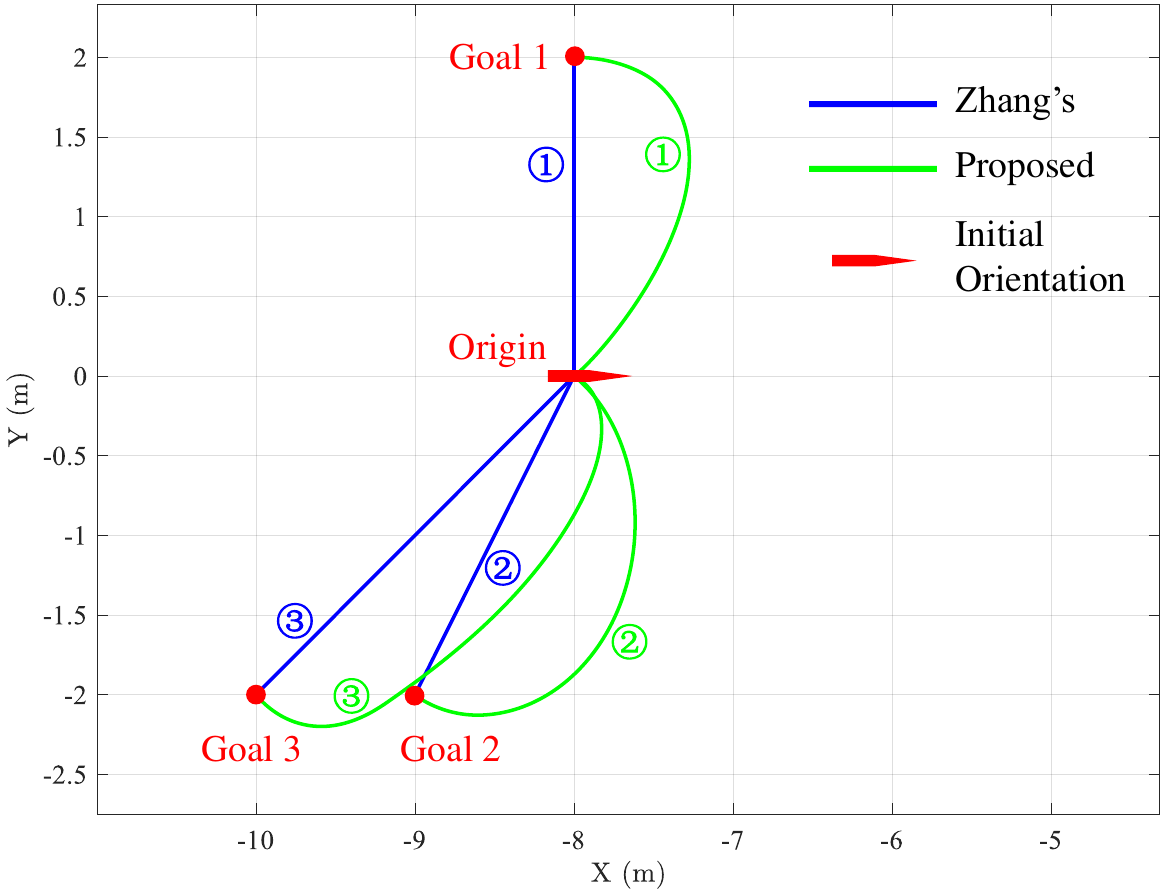}\\
    \caption{Comparsion of trajectory planning algorithm between the proposed and Zhang's \cite{c11} in simulation.}
    \label{Fig_traj_bench}
    \vspace{-0.3cm}
\end{figure} 

\subsection{Comparison of Terrestrial Trajectory Planning}

To demonstrate the adaptability of the proposed planner in crawling mode of HyFCQs, we compare it with Zhang's method \cite{c11}, which lacks consideration for nonholonomic constraints in path-searching. Firstly, a simulation experiment is implemented by the Robot Operating System (ROS). We publish a series of goals which are specially arranged to create scenarios involving nonholonomic constraints. As shown in Fig. \ref{Fig_traj_bench}, the proposed trajectory starts from the initial orientation and connects to the goal by a steering curvature, in contrast to Zhang's trajectory, which follows a straight line from the start point to the goal. This is common in practice, for example, a vehicle needs to turn around to reach a lateral or rear goal. 

\begin{figure}[h]
    \centering
    \setlength{\abovecaptionskip}{0.0cm}
    \includegraphics[width=0.48\textwidth]{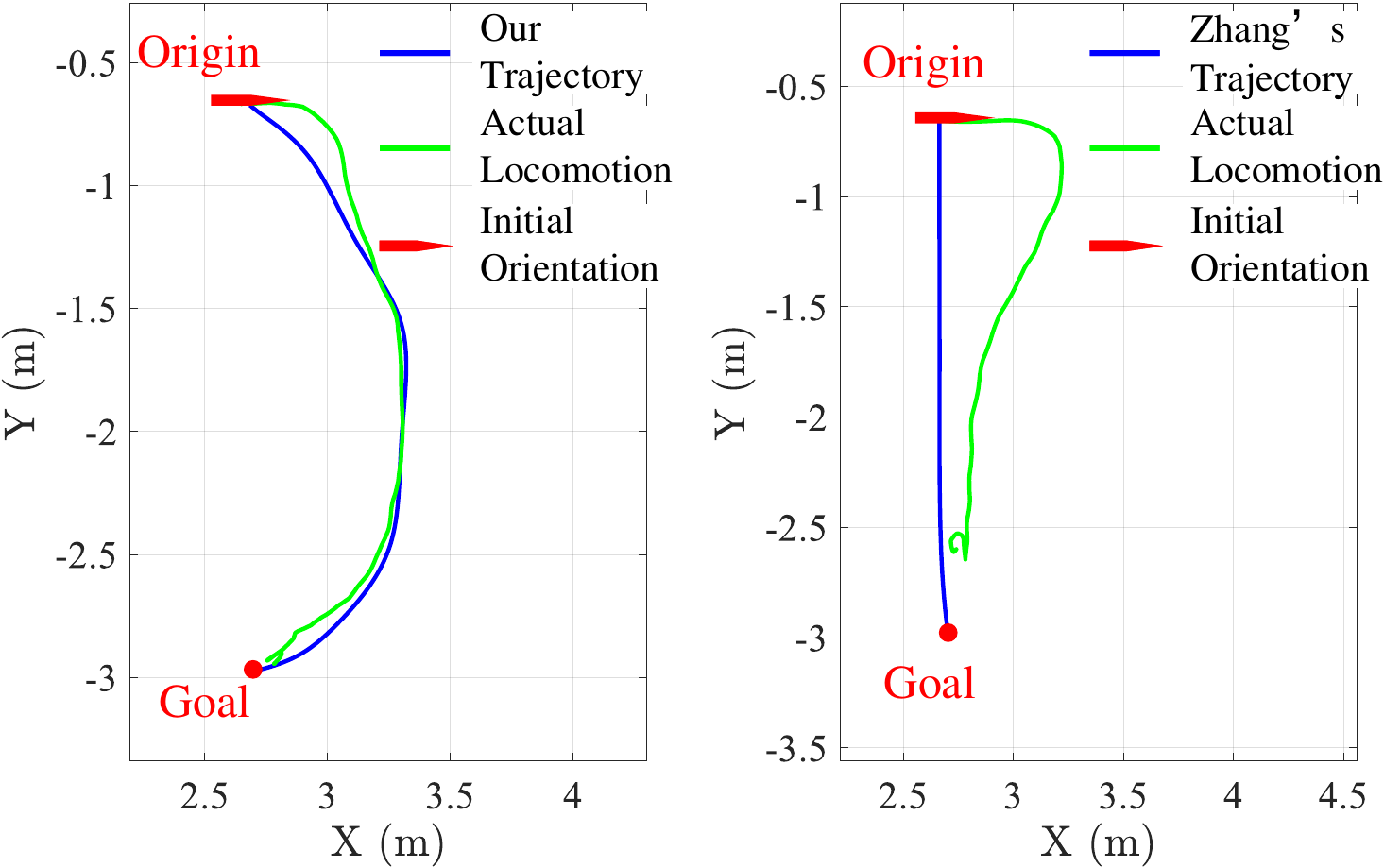}\\
    \caption{The Left) and Right) figures respectively represent our and Zhang's \cite{c11} trajectory planning in real-world experiments. Based on the tracking performance, our trajectory closely aligns with the crawling motion of HyFCQs in nonholonomic conditions.}
    \label{Fig_traj_real_bench}
    \vspace{-0.2cm}
\end{figure}

To verify that our trajectory meets the HyFCQ crawling motion requirements in the real world, a comparison of planning performance is made between our method and Zhang's under different velocity and acceleration limitations in nonholonomic conditions. The terrestrial tracking controller outlined in Sect \ref{section_tracking} is utilized for trajectory tracking with both planning methods. A lateral goal is set, and trajectories along with the locomotion path of the quadrotor are recorded. One experiment performance is shown in Fig. \ref{Fig_traj_real_bench}, in which the maximum velocity and acceleration of the trajectories are limited to 1.0 \(m/s\) and 0.8 \(m/s^2\) respectively. An evident observation from this figure is that our trajectory, in contrast to Zhang's, restricts changes in yaw angle, thus avoiding generating excessive steering motion over short distances. Table \ref{position_error_comparision} presents the average position error \(E_{ap}\) and maximum position error \(E_{mp}\) of the experiments. Where \(v_m\) and \(a_m\) represent maximum velocity and acceleration respectively. Zhang's \(E_{ap}\) is approximately 4.4 times greater than ours, while the \(E_{mp}\) is about 3.4 times greater than ours. These results suggest that Zhang's trajectories are kinodynamic infeasible for HyFCQs, whereas ours are feasible.

\begin{table}[t]
    \renewcommand{\arraystretch}{1.3}
    \caption{Position Error Comparision}
    \label{position_error_comparision}
    \centering
    \begin{tabular}{|c|c|c|c|c|}
        \hline
        Method & \(v_m (m/s)\) & \(a_m (m/s^2)\) & \(E_{ap}(m)\) & \(E_{mp}(m)\) \\
        \hline
        Proposed & \multirow{2}{*}{1.0} & \multirow{2}{*}{0.8} & \textbf{0.077} & \textbf{0.156} \\
        \cline{1-1} \cline{4-5}
        Zhang's \cite{c11} & & & 0.347 & 0.595 \\
        \hline
        Proposed & \multirow{2}{*}{1.2} & \multirow{2}{*}{1.0} & \textbf{0.080} & \textbf{0.174} \\
        \cline{1-1} \cline{4-5}
        Zhang's \cite{c11} & & & 0.354 & 0.598 \\
        \hline
    \end{tabular}
    \vspace{-0.3cm}
\end{table}

\begin{figure}[h]
    \centering
    \setlength{\abovecaptionskip}{0.0cm}
    \includegraphics[width=0.49\textwidth]{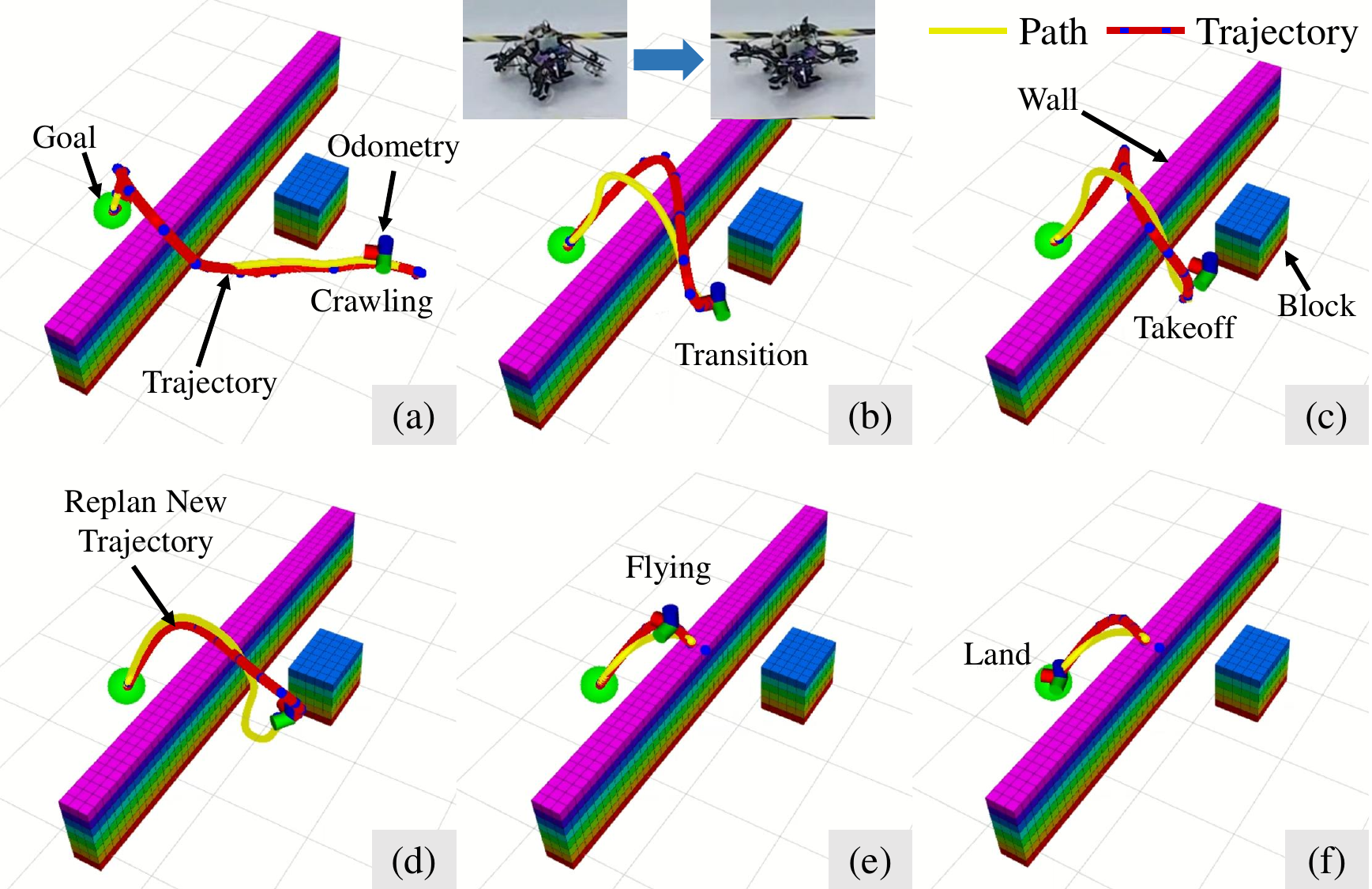}\\
    \caption{Autonomous navigation of the quadrotor involves bypassing a block and flying over a wall. The paths are generated through our search method and subsequently optimized into trajectories. The odometry data, represented by the red, green, and blue axes for the x, y, and z dimensions respectively, includes the current position and orientation.}
    \label{Fig_first_hy_exp}
    \vspace{-0.2cm}
\end{figure} 

\subsection{Hybrid Terrestrial-Aerial Navigation}
To validate the navigation capabilities of the hybrid trajectory planning and tracking system, two testing scenarios are established. Each scenario features insurmountable barriers, forcing the quadrotor to transition to flying mode to bypass these obstacles. The obstacle point cloud is pre-established, and the state observation data is obtained through visual inertial odometry (VIO) \cite{c15}. In both experiments, the maximum velocity is 1.0 \(m/s\), and the maximum acceleration is 0.8 \(m/s^2\). In the first scenario, a block and a long wall are positioned in front of the quadrotor, with the target point situated behind the wall. As shown in Fig. \ref{Fig_first_hy_exp}, the hybrid terrestrial-aerial trajectory that is generated allows the quadrotor to circumvent the block and surmount the wall. The quadrotor crawls along this trajectory until it reaches the transition point where terrestrial and aerial paths intersect. Then, HyFCQ takes off and ascends to a preset altitude. After that, the planner replans a new trajectory from the current position to the endpoint. The quadrotor then flies along this trajectory and finally lands at the goal. 

\begin{figure}[t]
    \centering
    \setlength{\abovecaptionskip}{0.0cm}
    \includegraphics[width=0.5\textwidth]{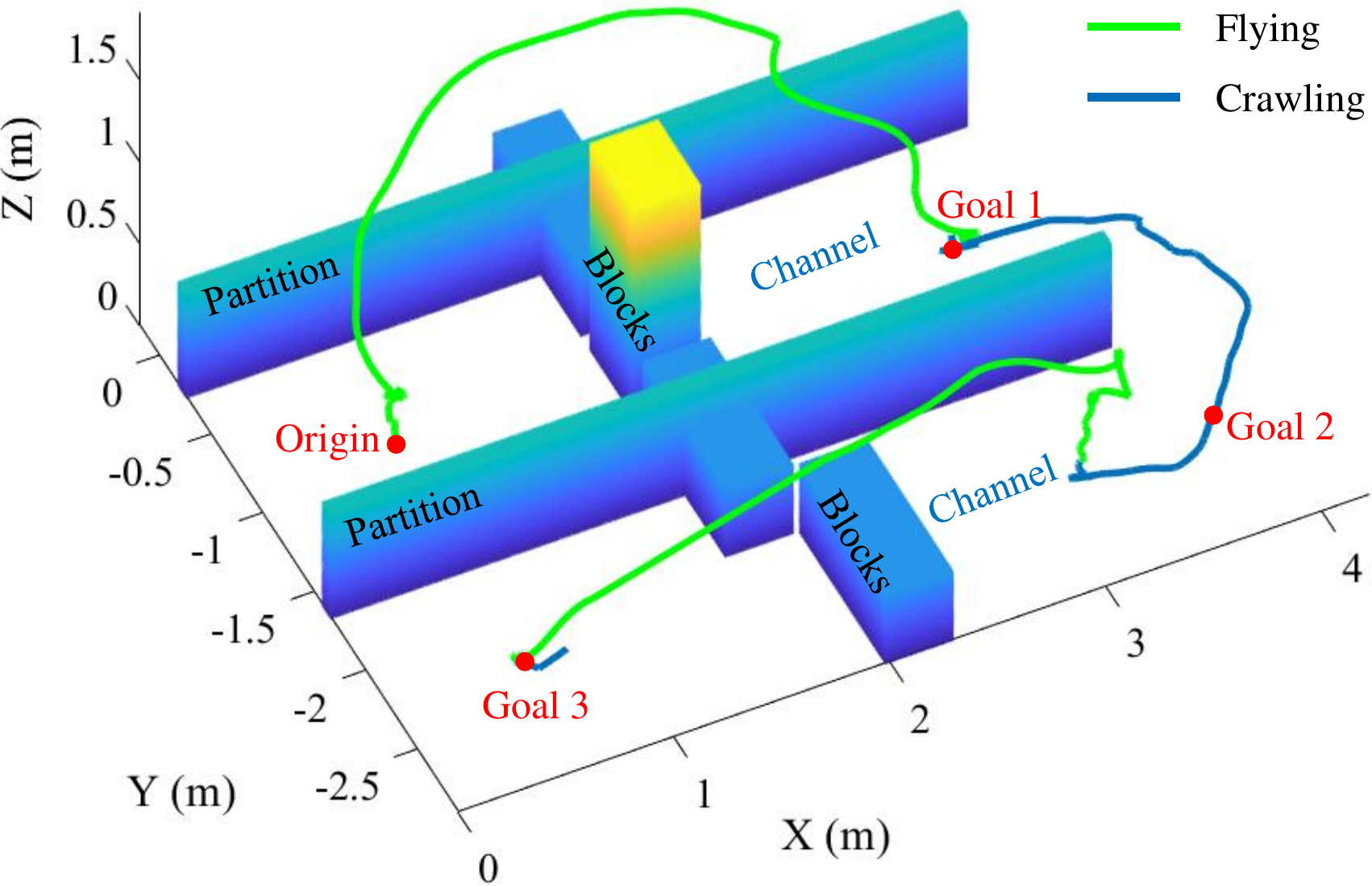}\\
    \caption{The hybrid terrestrial-aerial navigation performance in the second experiment. }
    \label{Fig_second_hy_exp}
    \vspace{-0.3cm}
\end{figure} 
\begin{figure}[t]
    \centering
    \setlength{\abovecaptionskip}{0.0cm}
    \includegraphics[width=0.49\textwidth]{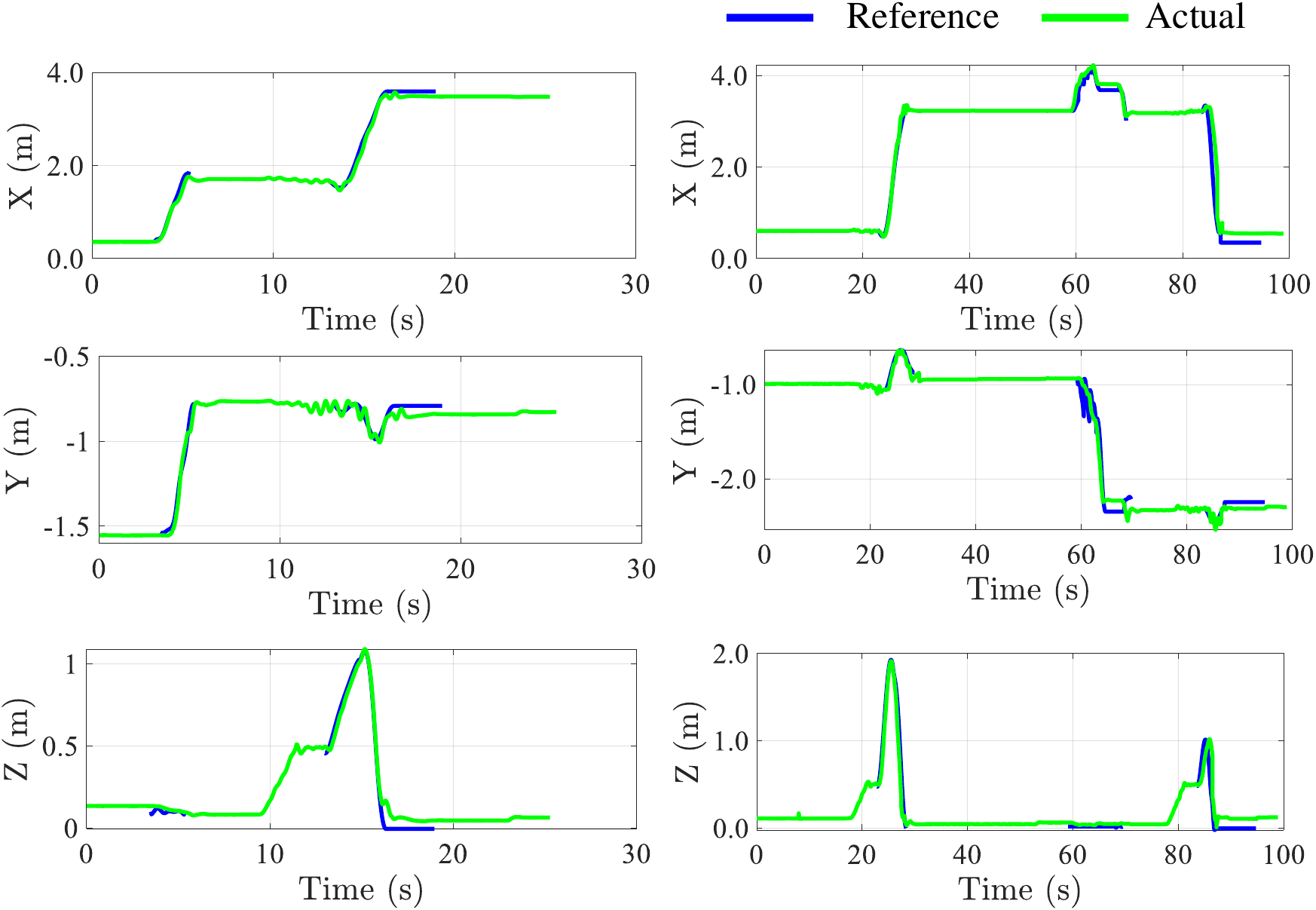}\\
    \caption{Actual and reference trajectories in the X, Y, and Z dimensions of the two navigation experiments. The Left) figure displays results from the first experiment, while the Right) figure shows results from the second experiment. The interruptions in the reference trajectory are due to the drone undergoing transformations, taking off, or landing, phases during which the trajectory planner does not engage in navigation tasks.}
    \label{Fig_hybrid_tracking}
    \vspace{-0.3cm}
\end{figure} 

In the second scenario, the experimental site is partitioned into two channel spaces, and numerous blocks are placed within these channels. As shown in Fig. \ref{Fig_second_hy_exp}, several goals are then published, allowing autonomous navigation through the obstacles. The navigation process of the quadrotor mirrors that of the first experiment, ultimately resulting in the quadrotor autonomously traversing the entire scene. 

Fig. \ref{Fig_hybrid_tracking} presents the trajectory tracking results of the two hybrid terrestrial-aerial navigation experiments, comparing the actual and reference trajectories in the X, Y, and Z dimensions. The average tracking errors for the first and second experiments are 0.102 \(m\) and 0.111 \(m\), respectively. The outcomes of these experiments validate the feasibility of the proposed trajectory planning and tracking method.



\section{CONCLUSION}

In this study, we propose a trajectory planning and tracking framework specifically designed to accommodate the unique motion characteristics of HyFCQs. To optimize the terrestrial path-searching of the hybrid motion planner, we incorporate crawling refinement to generate dynamically feasible trajectories. A novel trajectory tracking method optimizes autonomous navigation by re-planning trajectory and remapping control inputs for both crawling and flying modes. Furthermore, we conduct multiple simulations and real-world experiments to validate the effectiveness of the navigation system.

In the future, we plan to incorporate structural deformation constraints into the trajectories, enhancing the speed and smoothness of HyFCQs' motion.

\addtolength{\textheight}{-12cm}   





\end{document}